\pdfoutput=1
\documentclass[11pt]{article}

\usepackage[]{ACL2023}

\usepackage{times}
\usepackage{latexsym}
\usepackage{multirow} % for multiple rows
\usepackage{graphicx} % for scalebox

\usepackage[T1]{fontenc}
\usepackage[utf8]{inputenc}

\usepackage{microtype}

\usepackage{inconsolata}

\title{Language Variety Identification with True Labels}

\author{Marcos Zampieri\textsuperscript{1}, Kai North\textsuperscript{1}, Tommi Jauhiainen\textsuperscript{2}, Mariano Felice\textsuperscript{3} \\ {\bf
 Neha Kumari\textsuperscript{4}, Nishant Nair\textsuperscript{4}, Yash Bangera\textsuperscript{4}}
\vspace{2mm} \\
  \textsuperscript{1}George Mason University, USA, \textsuperscript{2}University of Helsinki, Finland \\
    \textsuperscript{3}Cambridge University, UK, \textsuperscript{4}Rochester Institute of Technology, USA \vspace{2mm} \\
  \texttt{mazgla@rit.edu} \\
}

\begin{document}
\maketitle
\begin{abstract}
Language identification is an important first step in many IR and NLP applications. Most publicly available language identification datasets, however, are compiled under the assumption that the gold label of each instance is determined by where texts are retrieved from. Research has shown that this is a problematic assumption, particularly in the case of very similar languages (e.g., Croatian and Serbian) and national language varieties (e.g., Brazilian and European Portuguese), where texts may contain no distinctive marker of the particular language or variety. To overcome this important limitation, this paper presents DSL True Labels (DSL-TL), the first human-annotated multilingual dataset for language variety identification. DSL-TL contains a total of 12,900 instances in Portuguese, split between European Portuguese and Brazilian Portuguese; Spanish, split between Argentine Spanish and Castilian Spanish; and English, split between American English and British English. We trained multiple models to discriminate between these language varieties, and we present the results in detail. The data and models presented in this paper provide a reliable benchmark toward the development of robust and fairer language variety identification systems. We make DSL-TL freely available to the research community. 
\end{abstract}

\section{Introduction}
\label{introduction}

Language identification is the task of automatically identifying the language of a given text or document \cite{jauhiainen2019survey}. The task is a vital pre-processing step integrated into many IR and NLP applications. Language identification is commonly modeled as a supervised text classification task where the archetypal language identification system typically follows these four main steps \cite{Lui2014GeneralizedLI}:

\begin{enumerate}
  \item[(a)]  Representation: selects a text representation (e.g., characters, words, or a combination of the two);
  \item[(b)]  Language Modelling: derives a model from texts for each language;
  \item[(c)]  Classification: defines a function that best represents the similarity between a text and each language model;
  \item[(d)]  Prediction: computes the highest-scoring model to determine the language of the given text.
\end{enumerate}

\noindent In the early 2000s, language identification was widely considered as a solved task \cite{mcnamee2005language} since character n-gram language models achieve perfect performance on discriminating between sets of dissimilar languages (e.g., Arabic, English, Finish, and Japanese) in standard contemporary texts (e.g., newspaper texts). Renewed interest in the task has emerged in the last decade with more challenging scenarios of particular interest to IR applications. This includes identifying the language of very short non-standard texts from user-generated content (e.g., microblogs) \cite{tromp2}, and web queries \cite{anand1,ceylan1}. Other challenges to state-of-the-art language identification systems arise from linguistic phenomena such as code-mixing and code-switching, where two or more languages are mixed in texts or social media posts \cite{solorio-etal-2014-overview,molina1}. 

Discriminating between very similar languages, dialects, and national varieties of the same language is another important, challenging language identification scenario that has been addressed by several studies \cite{tiedemann-ljubesic-2012-efficient,lui-cook-2013-classifying,bouamor2019madar}. In this scenario, systems need to model fine distinctions between a set of closely-related languages (e.g., Bulgarian and Macedonian), dialects (e.g., the different dialects of Arabic), or national varieties of the same language (e.g., Brazilian and European Portuguese) to accurately discriminate between them. This challenge has been the main topic of the workshop series on NLP for Similar Languages, Varieties, and Dialects (VarDial) \cite{vardial2020report, vardial2021report, aepli-etal-2022-findings} and their associated benchmark competitions which are organized yearly since 2014. The VarDial competitions have been providing the community with multiple datasets containing a wide variety of languages and dialects, helping to establish important language identification benchmarks. 

As discussed in Section \ref{related_work}, the main limitation of the datasets collected for VarDial and similar competitions is that the gold labels for each instance are not obtained through human annotation. The most widely used one, the DSL Corpus Collection (DSLCC) \cite{tan:2014:BUCC}, for example, contains news texts retrieved from multiple newspaper websites considering the domain of the website as a proxy for language variety. For example, all content retrieved from news websites hosted in country-specific domains such as .br and .pt domains is labeled as Brazilian and European Portuguese, respectively. While this is a straightforward assumption that results in a high number of accurate gold labels, this assumption has proved to be problematic in cases of republications of articles in different countries, particularly for languages that are widely spoken throughout the world, most notably English \cite{zampieri:2014:VarDial}. Furthermore, multiple studies \cite{acs-gradgyenge-rodriguesderezendeoliveira:2015:LT4VarDial,goutte2016discriminating} have evaluated native speakers' performance in identifying language varieties using the DSLCC concluding that many instances do not include any marker that allows humans to discriminate between varieties. 

To address this limitation, in this paper, we introduce DSL True Labels (DSL-TL), the first human-annotated language variety identification dataset. To the best of our knowledge, no manually annotated dataset with true labels is available for language variety identification or language identification in general, and ours fills this gap. We collect instances available in the DSLCC and in other news corpora and gather multiple human judgments for each instance through a crowdsourcing platform. Finally, we train and evaluate multiple machine-learning models on this new dataset. 

The contributions of this paper are the following:

\begin{enumerate}
    \item A novel problem formulation for language variety identification and language identification in general. 
    \item The release of DSL-TL, the first human-annotated language identification dataset.\footnote{\url{https://github.com/LanguageTechnologyLab/DSL-TL}}
    \item An evaluation of multiple language identification models on this new dataset.
\end{enumerate}

\noindent The remainder of this paper is organized as follows. Section \ref{related_work} discusses prior research in language variety identification, including the VarDial competitions and available datasets. Section \ref{dataset} details the steps taken in the construction DSL-TL dataset from data collection to annotation. Section \ref{models} describes the language identification models used in our experiments, while Section \ref{sec:results} presents their results on the new DSL-TL dataset. Finally, Section \ref{conclusion} summarizes our research and discusses avenues for future work.

\section{Related Work} 
\label{related_work}

As discussed in a recent survey \cite{jauhiainen2019survey}, several language identification studies have reported achieving near-perfect performances in a variety of scenarios. \citet{Ljubesic2014DiscriminatingBV} trained a selection of traditional machine learning classifiers to discriminate between social media posts (tweets) written in four related languages: Bosnian, Croatian, Montenegrin, and Serbian. Their best model, being a Gaussian naive Bayes (GNB) classifier, achieved an accuracy of 97.1\%. \citet{Martadinata_etal_2016} used a Markov model to identify which extracts were taken from Wikipedia articles in Indonesian, Javanese, Sundanese, or Minang‐kabau. Their model achieved an accuracy of 95.75\%. \citet{li-etal-2018-whats} trained a convolutional neural network (CNN) on multiple datasets containing 97 languages. Their model consistently achieved performance of over 95\% accuracy when training was conducted across several datasets. 

Language variety identification systems that discriminate between varieties of the same language, however, achieve more varied performances as discussed in the VarDial shared task reports \cite{vardial2020report, vardial2021report, aepli-etal-2022-findings}. Since 2014, the VarDial workshop has hosted several shared tasks for language variety identification, as discussed next.

\subsection{VarDial Shared Tasks}
\label{VarDial}

The Discriminating between Similar Languages (DSL) shared task at VarDial-2014 \cite{zampieri:2014:VarDial}, saw eight teams produce systems for distinguishing between similar languages and language varieties of several language groups. The best-performing model used a probabilistic model similar to a Naive Bayes classifier combined with several SVMs. They reported an accuracy of 91\%  for differentiating between European and Brazilian Portuguese and an accuracy of 95.6\% for Castilian and Argentine Spanish \cite{goutte-leger-carpuat:2014:VarDial}. The same model achieved an accuracy of 52.2\% for differentiating between British and American English \cite{zampieri:2014:VarDial}. 

DSL continued in VarDial-2015 and 2016 \cite{zampieri:2015:LT4VarDial, dsl2016}. Both iterations of this shared task expanded upon the original dataset. The 2015 edition added several additional languages and removed named entities to determine their influence on performance \cite{zampieri:2015:LT4VarDial}. The 2016 edition included varieties of French and challenged 18 teams with Arabic dialect identification \cite{dsl2016}. The highest performing systems \cite{malmasi-zampieri:2016:VarDial3, ionescu-popescu:2016:VarDial3, eldesouki-EtAl:2016:VarDial3, adouane-semmar-johansson:2016:VarDial3} achieved accuracies ranging from 49.7\% and 51.2\% when differentiating between Egyptian, Gulf, Levantine, Modern Standard, and Maghreb dialects. The two best-performing systems were an ensemble or a single SVM(s) trained on character and word-level n-grams \cite{malmasi-zampieri:2016:VarDial3, eldesouki-EtAl:2016:VarDial3}.

Since 2017 VarDial continued to host shared tasks for identifying other language varieties \cite{vardial2017report, vardial2018report,vardial2019report, vardial2020report, vardial2021report, aepli-etal-2022-findings}. Performances on these shared tasks were consistent with that of 2014 to 2016, with SVMs and models trained on character and word-level n-grams often outperforming other approaches. An ensemble or a single SVM(s) trained on character n-grams achieved the highest performance for language variety identification for German in 2017 with an F1 of 0.662 \cite{malmasi-zampieri:2017:VarDial1}, for Dutch and Flemish in 2018 with an F1 of 0.660 \cite{ccoltekin-rama:2017:VarDial}, and for Romanian in 2020 with an F1 of 0.787 \cite{coltekin-2020-dialect}. Naive Bayes trained on character n-grams also reported the highest F1s of 0.908 for Chinese in 2019 \cite{jauhiainen-etal-2019-discriminating}, 0.777 for Romanian in 2021 \cite{jauhiainen2021naive}, and 0.9 for Italian in 2022 \cite{jauhiainen-etal-2022-italian}. Language identification is, therefore, far from being a solved task, with performances varying greatly between groups of dialects and language varieties. Dataset quality and the similarity between language varieties are responsible for such varied performances.

%  European (1,317) and Brazilian Portuguese (3,023), Castilian (2,131), and Argentine Spanish (1,211), along with  British (1,081)
 \begin{table*}[!ht]
% \begin{adjustbox}{width=\columnwidth,center}
\centering
\scalebox{0.92}{\begin{tabular}{l|ccc|c}
% \hline
% \multicolumn{1}{l}{} & \multicolumn{3}{c}{\textbf{Class Split}} & \multicolumn{1}{}{} \\
    \hline
     \textbf{Language} & \textbf{Variety A} &  \textbf{Variety B} &  \textbf{Both or Neither} & \bf Total \\ 
     \hline
     Portuguese & 1,317 (pt-PT) & 3,023 (pt-BR) & 613 (pt) & 4,953\\
     Spanish & 2,131 (es-ES) & 1,211 (es-AR) & 1,605 (es) & 4,947\\
     English & 1,081 (en-GB) & 1,540 (en-US) & 379 (en) & 3,000\\
 \hline
\bf Total & & & & \bf 12,900 \\
 \hline
\end{tabular}}
% \end{adjustbox}
 \caption{\label{dataset_table1} DSL-TL's class splits and the total number of instances.}
\end{table*}

\subsection{Available Datasets}

The datasets used in the VarDial shared tasks, as well as other similar shared tasks \cite{zubiaga2016tweetlid,rangel2017overview}, contain thousands of sentences in groups of languages or dialects sampled mostly from local newspapers and social media. Examples include Portuguese, Spanish, and English \cite{zampieri:2014:VarDial, zampieri:2015:LT4VarDial}, Arabic \cite{dsl2016}, Chinese \cite{vardial2019report}, Romanian \cite{vardial2020report}, and Italian \cite{aepli-etal-2022-findings}. However, as discussed in the introduction, these datasets consist of instances assigned with a ground truth label determined by where the text was published (e.g., UK, USA, etc.). Each sentence within these datasets is, therefore, either, for example, American or British English, and only one of these labels is considered correct according to the gold labels included in these datasets.

The problem with formulating automatic language identification in this way is that many sentences do not necessarily belong to a single language variety \cite{goutte2016discriminating}. This is true for varieties of English and varieties of other languages. The DSL dataset from 2014 \cite{zampieri:2014:VarDial}, for example, contained instances with incorrect ground truth labels. Articles containing features characteristic of British English were published in American newspapers and vice-versa, resulting in mislabeled data. \citet{goutte2016discriminating} went on to show that human annotators were unable to achieve competitive performances on the DSL 2014 and 2015 datasets \cite{zampieri:2014:VarDial, zampieri:2015:LT4VarDial}. On average, accuracies achieved by single human annotators were just over 50\%, with performances varying between languages. Relying on the source of extracts or on an individual annotator to determine binary ground truth labels is, therefore, problematic. We address this limitation with DSL-TL by collecting the first human-annotated language identification dataset containing multiple annotations per instance. 

\section{The Dataset: DSL-TL} \label{dataset}

\subsection{Rationale and Motivation}
\label{motivation}

The main limitation of the datasets used in the aforementioned benchmark competitions is that each instance (a sentence or a paragraph) contains only one ground truth label, which is assigned depending on the country where the text was published (e.g., UK, USA, etc.). Therefore, each sentence in the datasets is either, say, American or British English, and only one of the labels is considered correct when evaluating the language identification system. The problem with this task formulation is that, as demonstrated in previous research \cite{goutte2016discriminating}, many sentences are simply impossible to be identified because they may belong to multiple language varieties. For example, not all sentences published in a British newspaper contain features that are exclusive to British English, such as spelling conventions (e.g., analyze, neighbor) or lexical choices (e.g., trousers, rubbish) that would make it possible for an English native speaker to recognize a sentence as British. The same is true for other English varieties and varieties of other languages. \citet{goutte2016discriminating} showed that native speakers perform very poorly in identifying the language variety when only one label needs to be assigned to each text.

As human performance is often below chance in this task, it is unfair to expect that automatic systems will ever be able to achieve 100\% performance when discriminating between national varieties of the same language. To cope with this important limitation, we introduce the use of true labels in language variety identification. The true labels are designed to capture the presence or absence of variety-specific features in each sentence by collecting and aggregating multiple human judgments per data point using Amazon Mechanical Turk (AMT). The annotators were displayed with sentences from the dataset, and they were asked to assign one of the following three labels to a given sentence.

\begin{itemize}
    \item variety X when at least one feature of the variety X is present in the instance;
    \item variety Y when at least one feature of the variety Y is present in the instance;
    \item both/neither when no or the same number of features of the variety X and Y is present in the instance.
\end{itemize}

\noindent The annotators were paid between 3 and 5 cents of US dollar per annotation. More detail on the data collection and annotation is provided next.

\subsection{Data}
 
DSL-TL contains 12,900 instances split between several language varieties, as shown in Table~\ref{dataset_table1}. These instances vary between 1 to 3 sentences in length. They consist of short extracts taken from newspaper articles. The English articles have been sourced from a collection of news articles made available by \citet{zellers2019defending} - henceforth True News - while the Portuguese and Spanish articles have been sourced from the DSLCC \cite{tan:2014:BUCC}. Both datasets feature data retrieved from multiple newspapers from each country. We randomly selected instances from the original datasets with an even split between each language variety, being a 2,500/2,500 split for Portuguese and Spanish varieties and a 1,500/1,500 split for English varieties. The final 12,900 instances in DSL-TL have been randomly split into training, development, and testing partitions in a 70\%, 20\%, 10\% split as shown in Table \ref{train_test_splits}. Finally, example instances from DSL-TL are provided in Table \ref{example_instance}.

 \begin{table}[!ht]
% \begin{adjustbox}{width=\columnwidth,center}
\centering
\scalebox{0.92}{\begin{tabular}{l|ccc|c}
% \hline
% \multicolumn{1}{l}{} & \multicolumn{3}{c}{\textbf{Class Split}} & \multicolumn{1}{}{} \\
    \hline
     \textbf{Variety} & \textbf{Train} &  \textbf{Dev} &  \textbf{Test} & \bf Total \\ 
     \hline
     Portuguese & 3,467 & 991 & 495 & 4,953\\
     Spanish & 3,467 & 985 & 495 & 4,947\\  % missing 3
     English & 2097 & 603 & 300 & 3,000\\ % missing 4
 \hline
 Total & & & & 12,900 \\
 \hline
\end{tabular}}
% \end{adjustbox}
 \caption{\label{train_test_splits} DSL-TL's train, dev, and test splits are 70/20/10\% of the total number of instances, respectively.}
\end{table}

\begin{table*}[!ht]
\centering
\scalebox{0.8}{\begin{tabular}{l|p{12cm}|c|c|c}
    \hline
      &  &  \multicolumn{2}{c|}{\textbf{Gold Label}} & \\
    \hline
        \textbf{Language} & \textbf{Sentence} &  \textbf{Old} &  \textbf{DSL-TL} & \textbf{Markers} \\
      \hline
    \multirow{3}{*}{Portuguese} & desde a \textbf{cracolândia} até as grandes mansões que existem à beira-mar... & pt-BR & pt-BR  & cracolândia\\
    
    &O \textbf{reajuste} solicitado pela Celpe, empresa do Grupo Neoenergia, previa o efeito médio de 8,67\%. 
 & pt-BR & pt-BR & reajuste \\

     &O Esta \textbf{equipa} do Athletic Bilbau é muito diferente da que, em 1976-77, jogou e perdeu a final.
 & pt-PT & pt-PT & equipa \\
    
    \hline
    \multirow{3}{*}{Spanish} &Le dejé la llave a un \textbf{baqueano} y debo volver en poco tiempo, porque en abril las condiciones... & es-AR & es-AR & baqueano \\

    & Aún así, la alimentación sana consigue más \textbf{adeptos} cada día gracias... restaurantes verdes. & es-ES & es-ES & adeptos \\
    
    & Estas irregularidades fueron planteadas por legisladores del \textbf{PSOE} y, sobre todo, de Izquierda... & es-AR & es-ES & PSOE \\
    
    \hline
   \multirow{3}{*}{English} & A rose Hill funeral director who prides herself on delivering the perfect \textbf{personalised} send-off & en-GB & en-GB & personalised\\
   
    & Ten symbolic silhouettes are on display around the \textbf{county} as part of the... campaign. & en-GB & both/neither & county\\

    & It seems very un-Saintslike, almost unnatural, to go into the last day of the season with nothing.. & en-GB & en-GB & saintslike\\

    \hline
\end{tabular}}
 \caption{Example instances in English, Portuguese, and Spanish from the DSL-TL. The `old label' represents the original dataset label along with the new label at the DSL-TL corpus. The linguistic markers identified by the annotators are provided in bold. Only a snapshot of these instances is shown.} \label{example_instance}
%  {\label{tab:table-name}target words are in bold.}
\end{table*}

\subsection{Annotation}
\label{annotation}

The annotators were crowd-sourced using AMT. They were based in the six countries where the language varieties were spoken, namely Argentina, Brazil, Portugal, Spain, United Kingdom, and the United States. The annotators were requested to label instances in their own native or non-native language variety. They labeled instances as being either European (pt-PT) or Brazilian Portuguese (pt-BR), Castilian (es-ES) or Argentine Spanish (es-AR), and British (en-GB) or American English (en-US). Label distributions are shown in Table \ref{dataset_table1}.

We asked annotators to label each instance with what they believed to be the most representative variety label. They were presented with three choices: (1). language variety A, (2). language variety B, or (3). both or neither. We initially collected three annotations for each of the 12,900 instances in the dataset. We considered the gold label correct in cases in which the three annotators agreed on the same label or when two annotators agreed with the original gold label - the DSLCC for Spanish and Portuguese and True News for English. This resulted in 6,426 instances annotated by three annotators. For the remaining 6,474 instances, we collected two additional human annotations targeting an agreement of at least three annotators agreeing on the label or two annotators agreeing with the original dataset's gold label. 

Finally, the annotators were also asked to identify the linguistic markers or named entities that influenced their decision. From the total 12,900 instances, 3,386 instances were provided with linguistic markers. The number of markers for each language is 270 for English, 2,378 for Spanish, and 738 for Portuguese.

\section{Models}
\label{models}

\begin{table*}[!ht]
\centering
\scalebox{0.8}{\begin{tabular}{c|c|c|c}
    \hline
         &\textbf{mBERT} & \textbf{XLM-R}  & \textbf{XLM-R-LD} \\
      \hline
     type & BERT-base & RoBERTa-base & RoBERTa-base \\
     corpus & Wikipedia & CC data &  LID \\
     size & 3.3B (102 lang.) & 2.5TB (100 lang.) & 70k (20 lang.)\\
     \#layers & 12 & 12 & 12  \\
     \#heads  & 12 & 16 & 16  \\
     \#lay.size  & 768 & 768 & 768 \\
     \#para  & 110M & 250M & 250M \\
     \hline
\end{tabular}}
 \caption{Comparison of mBERT, XLM-R, and XLM-R-LD models. Lang is short for languages. CC data refers to CommonCrawl data. %\cite{wagner-filho-etal-2018-brwac}
 }\label{model_parameters}
%  {\label{tab:table-name}target words are in bold.}
\end{table*}

We trained classic machine learning models as well as transformer-based models on the DSL-TL corpus, as presented in the next sections. As discussed by \citet{medvedeva-kroon-plank:2017:VarDial} and \citet{jauhiainen2019survey}, deep learning models have not shown to clearly outperform traditional machine learning models in language identification, so we take this opportunity to test 
methods from different machine learning paradigms. 

The models were evaluated in two tracks. 

\begin{itemize}
    \item {\bf Track one} contains nine labels - the six language varieties plus the both or none class for each language. 
    \item {\bf Track two} contains six labels - only the six language varieties
\end{itemize}

\subsection{Naive Bayes}

We describe experiments using the newest version of the Naive Bayes system previously used by \citet{jauhiainen:2022:nadi} and \citet{jauhiainen-etal-2022-italian}. For each language pair, we used the common instances as a third "language" in a usual classification setup on track one. 

%Additionally, we added the common labeled training data to both languages training set. When optimizing the system parameters the development data was used in three non-overlapping sets, but for the final identification on the test data, the common labeled development instances were added to the training data for both languages in addition to their own development instances.

On our first run on the test data with the optimized parameters and the development data added as additional training data, the macro F1 was 0.505 for track one. Only character trigrams were used as they performed best on the development data. Table~\ref{tab:testing1} shows the statistics for individual language varieties for track one. The macro F1 is clearly affected by the low F1 scores of the common instances. For track two, we modified the system to ignore the common instances when evaluating during optimization. The common instances in the training and the development sets were added to both varieties in each language. This time the optimal character n-gram range was from two to five. The system attained a macro F1 score of 0.794, which also outperformed the deep learning systems with pre-trained language models.

For track two, we also experimented with the open variety by adding the training data from DSLCC v1.0 corpus. Adding the English instances from the first DSL shared task made the results worse on the development set, whereas adding the Portuguese and the Spanish instances improved the results. This is probably due to the fact that the results attained for the English varieties were already much higher than those of the best systems on DSL 2014. The resulting macro F-score for the open NB run was 0.803.

%Table~\ref{tab:testing2} shows similar statistics for track 2.

\subsection{Adaptive Naive Bayes}

In addition to the traditional Naive Bayes identifier, we used it with adaptive language models \cite{jauhiainen2019nle} in a similar manner to \citet{jauhiainen-etal-2022-italian} in the winning system of the ITDI shared task \cite{aepli-etal-2022-findings}. For the adaptive version of the Naive Bayes classifier, we use the same penalty modifier and character n-gram range as in the non-adaptive version. Using the development data, we optimize the number of splits used in adaptation as well as the number of learning epochs used. The number of splits indicates the size of the most confidently identified test data to be added after each identification run. The number of splits was chosen to be 512 with four epochs of adaptation. Table~\ref{tab:testing1} shows the results for track one. The macro averaged F1-score of 0.503 is slightly higher than that of 0.501, which was attained by the identical system without adaptation. On track two, a similar increase in performance was observed, with 0.799 attained by adaptive naive Bayes.
%On track two, the adaptive version was able to attain a macro F1 score of 0.7524, which is clearly higher than the 0.6957 of the corresponding non-adaptive version of the classifier. The results for individual languages on tracks two are shown in Table~\ref{tab:testingadap2}.

\subsection{Deep Learning Models}

We also experimented with several pre-trained large language models (LLMs). These LLMs were multilingual, and consisted of multilingual BERT\footnote{\url{https://huggingface.co/bert-base-multilingual-cased}}  (mBERT) \cite{devlin2019bert}, XLM-RoBERTa\footnote{\url{https://huggingface.co/xlm-roberta-base}}  (XLM-R) \cite{liu2019roberta}, and XLM-R-Language Detection\footnote{\url{https://huggingface.co/papluca/xlm-roberta-base-language-detection}} (XLM-R-LD). XLM-R-LD is a fine-tuned XLM-R model on the language identification dataset\footnote{\url{https://huggingface.co/datasets/papluca/language-identification}} (LID) containing 90k instances in 20 languages. These instances were taken from a range of sources, including Amazon reviews and SemEval tasks from 2012 to 2017. The three models were trained on train and dev sets with no bleed between sets (Table \ref{train_test_splits}). Train sets for English, Spanish, and Portuguese consisted of 2097, 3467, and 3467 instances, respectively. The dev sets contained 599 instances for English, 989 instances for Spanish, and 991 instances for Portuguese. Models were trained with a learning rate of 2e-5 over 5 epochs. Our models are summarized in Table \ref{model_parameters}.

% \subsection{mBERT}

% \subsection{XLM-R}

% \subsection{XLM-R-LD}

%The results of the optimization runs for both the adaptive and the non-adaptive Naive Bayes are found next.

\section{Results}
\label{sec:results}

In this section, we present the results obtained by all models in tracks one and two. Table \ref{tab:testing1} presents the results of the models on track one in terms of Precision, Recall, and F1-score, as well as the macro average for all languages. 

\begin{table}[!ht]
\centering
\footnotesize
\begin{tabular}{lllll}
\hline
\bf Variety & \bf Model & \bf Recall & \bf Prec. & \bf F1\\
\hline
 \multirow{5}{*}{\textbf{en-GB}} 
 & NB & 0.754 & 0.705 & 0.729 \\
 & ANB & 0.772 & 0.721 & 0.746 \\
 & mBERT & 0.760 &    0.807 &    0.783  \\
& XLM-R & 0.750  &   0.842   &  0.793 \\
& XLM-R-LD & 0.771 &    0.798  &   0.784 \\
\hline
 \multirow{5}{*}{\textbf{en-US}} 
  & NB & 0.750 & 0.848 & 0.796 \\
 & ANB & 0.731 & 0.851 & 0.786 \\
 &  mBERT  & 0.867 &    0.795  &   0.829   \\
& XLM-R & 0.829  &   0.776     &0.801 \\
& XLM-R-LD&  0.797 &    0.782  &   0.790\\
\hline
 \multirow{5}{*}{\textbf{en}} 
  & NB & 0.267 & 0.190 & 0.222 \\
 & ANB & 0.233 & 0.146 & 0.179 \\
 &   mBERT & 0.278  &   0.333  &   0.303   \\
&XLM-R  &  0.231   &  0.200   &  0.214   \\
&XLM-R-LD &  0.233 &0.233    & 0.233 \\
\hline
\multirow{5}{*}{\textbf{es-AR}} 
 & NB & 0.481 & 0.427 & 0.452 \\
 & ANB & 0.579 & 0.458 & 0.512 \\
&  mBERT  & 0.551 &    0.489   &  0.518  \\
&XLM-R &  0.551     &0.489     &0.518  \\
& XLM-R-LD & 0.511 &    0.519   &  0.515  \\
\hline
\multirow{5}{*}{\textbf{es-ES}} 
 & NB & 0.636 & 0.679 & 0.657 \\
 & ANB & 0.612 & 0.716 & 0.660 \\
&  mBERT  & 0.651  &   0.670   &  0.660  \\
&XLM-R  &  0.689   &  0.752    & 0.719 \\
& XLM-R-LD & 0.684&     0.694 &    0.689  \\
\hline
\multirow{5}{*}{\textbf{es}} 
 & NB & 0.327 & 0.336 & 0.331 \\
 & ANB & 0.34 & 0.351 & 0.345 \\
&   mBERT & 0.442   &  0.468  &   0.455  \\
&XLM-R  & 0.454  &   0.442     &0.448  \\
&XLM-R-LD & 0.444  &   0.429 &    0.436 \\
\hline
\multirow{5}{*}{\textbf{pt-BR}} 
 & NB & 0.662 & 0.762 & 0.708 \\
 & ANB & 0.609 & 0.795 & 0.689 \\
&  mBERT  & 0.718&     0.799  &   0.756   \\
&XLM-R  & 0.753&     0.786   &  0.769  \\
&XLM-R-LD & 0.739 &    0.796   &  0.767  \\
\hline
\multirow{5}{*}{\textbf{pt-PT}} 
 & NB & 0.533 & 0.442 & 0.483 \\
 & ANB & 0.555 & 0.442 & 0.492 \\
&  mBERT  & 0.459    & 0.496  &   0.477 \\
&XLM-R  & 0.492 &    0.657   &  0.562 \\
&XLM-R-LD &  0.488   &  0.613     &0.544  \\
\hline
\multirow{5}{*}{\textbf{pt}} 
 & NB & 0.136 & 0.118 & 0.126 \\
 & ANB & 0.153 & 0.100 & 0.121 \\
&  mBERT  & 0.214    & 0.051 &    0.082    \\
&XLM-R  & 0.000 &    0.000  &   0.000   \\
&XLM-R-LD & 0.000   &  0.000  &   0.000   \\
\hline
\multirow{5}{*}{\textbf{Macro}}  
 & NB & 0.505 & 0.501 & 0.501 \\
 & ANB & 0.509 & 0.509 & 0.503 \\
& mBERT & \bf 0.549     & \bf 0.545   &  \bf 0.540   \\
 &XLM-R  & 0.528  &   0.549  &   0.536  \\
 &XLM-R-LD& 0.519   &  0.541  &   0.529  \\
\hline
\end{tabular}
\caption{The scores for individual language varieties on the test set with Naive Bayes, Adaptative Naive Bayes, mBERT, XLM-R, and XLM-R-LD on track one in terms of Recall, Precision, and F1-score. Macro average is reported for average. The best average results are in bold.}
\label{tab:testing1}
\end{table}

\noindent In track one, we show that mBERT achieves the best results with a 0.540 average F1-score, followed by XLM-R with a 0.540 average F1-score. In terms of the performance for individual languages, we observe that all models obtained their best results for the two English varieties and, in particular, for en-US, with results as high as 0.829 F1-score obtained by the XLM-R model. This is somewhat surprising given that the English dataset is the smallest among the three languages. 

Finally, for all languages, the results obtained by all models for the `both or neither' class (en, es, and pt) were very low compared to the scores obtained for the varieties. This suggests that the class is very difficult to model due to the absence of any specific features. Previously released language identification datasets have not been manually annotated; they did not contain such a class. Therefore, the results on `both or neither' require further investigation.

\begin{figure}[!ht]
\centering
  \includegraphics[width=1.02\linewidth]{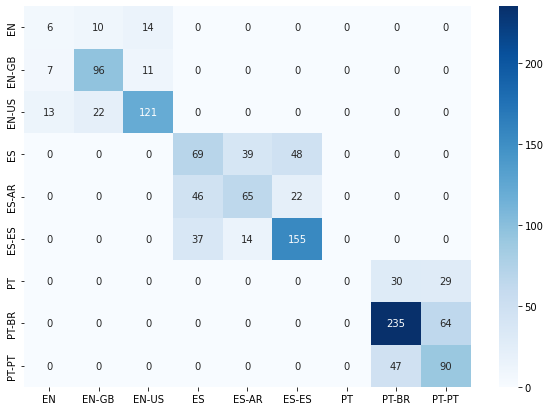}
  \caption{Confusion matrix showing the class predictions of mBERT in track one.}
  \label{confusion_matrix}
\end{figure}

To further understand the prediction of our best-performing model, mBERT, in Figure \ref{confusion_matrix} we plot a confusion matrix of our model in the track one setting. The confusion matrix shows that confusion does not occur outside the three labels of each language which is evidence of the high performance of the model in discriminating between different languages. We observe that the predictions for the `both or neither' Portuguese class behave differently than the classes for any of the other languages without any correct prediction.

\noindent In Table \ref{tab:testing2}, we present the results of all models in track two. We include the same five models as in track one plus a variation of Naive Bayes (NB, open) that has been trained using additional original data retrieved from the DSLCC. Unsurprisingly, the use of additional training data has boosted this system's performance and helped it achieve the best average score among all models with a 0.803 average F1-score. That said, we observed that in track two, the three Naive Bayes variations had outperformed the deep learning systems corroborating the findings of previous studies \cite{medvedeva-kroon-plank:2017:VarDial,jauhiainen2019survey}. Language identification is essentially a pattern-matching task rather than a semantic understanding one. We believe that this often favors relatively simpler character n-gram models when compared to more sophisticated text embedding-based representations. 

\begin{table}[!ht]
\centering
\footnotesize
\begin{tabular}{lllll}
\hline
\bf Variety & \bf Model & \bf Recall & \bf Prec. & \bf F1\\
\hline
 \multirow{3}{*}{\textbf{en-GB}} 
 & NB & 0.921 & 0.761 & 0.833 \\
 & NB, open & 0.877 & 0.758 & 0.813 \\
 & ANB & 0.93 & 0.752 & 0.831 \\
 & mBERT & 0.828  &  0.842    & 0.835 \\
& XLM-R & 0.795  &   0.921    & 0.854  \\
& XLM-R-LD & 0.802 &    0.851 &    0.826 \\
\hline
 \multirow{3}{*}{\textbf{en-US}} 
 & NB & 0.808 & 0.933 & 0.866 \\
 & NB, open & 0.821 & 0.901 & 0.859 \\
 & ANB & 0.795 & 0.939 & 0.861 \\
 &  mBERT  & 0.889 &    0.872 &    0.880    \\
& XLM-R & 0.935 &    0.827 &    0.878  \\
& XLM-R-LD&  0.886   &  0.846    & 0.866 \\
\hline
\multirow{3}{*}{\textbf{es-AR}} 
& NB & 0.857 & 0.726 & 0.786 \\
& NB, open & 0.789 & 0.789 & 0.789 \\
& ANB & 0.887 & 0.724 & 0.797 \\
&  mBERT  & 0.772    & 0.662  &   0.713     \\
&XLM-R &  0.750 &    0.654    & 0.699  \\
& XLM-R-LD&0.765   &  0.684   &  0.722   \\
\hline
\multirow{3}{*}{\textbf{es-ES}} 
& NB & 0.791 & 0.896 & 0.840 \\
& NB, open & 0.864 & 0.864 & 0.864 \\
& ANB & 0.782 & 0.915 & 0.843 \\
&  mBERT  & 0.800 &    0.874  &   0.835     \\
&XLM-R  &  0.794&     0.859    & 0.825 \\
& XLM-R-LD& 0.809   &  0.864   &  0.836    \\
\hline
\multirow{3}{*}{\textbf{pt-BR}} 
& NB & 0.716 & 0.873 & 0.787 \\
& NB, open & 0.696 & 0.924 & 0.794 \\
& ANB & 0.702 & 0.897 & 0.788 \\
&  mBERT  &  0.766  &   0.779    & 0.773  \\
&XLM-R  &  0.823 &    0.809   &  0.816  \\
&XLM-R-LD & 0.810   &  0.796  &   0.803   \\
\hline
\multirow{3}{*}{\textbf{pt-PT}} 
& NB & 0.774 & 0.564 & 0.652 \\
& NB, open & 0.876 & 0.58 & 0.698 \\
& ANB & 0.825 & 0.568 & 0.673 \\
&  mBERT  &  0.504    & 0.489 &    0.496\\
&XLM-R  & 0.599 &    0.620  &   0.609  \\
&XLM-R-LD &  0.570   &  0.591 &    0.581   \\
\hline
\multirow{3}{*}{\textbf{Macro}}  
& NB & 0.811 & 0.792 & 0.794 \\
& NB, open & \bf 0.820 & \bf 0.803 & \bf 0.803 \\
& ANB & 0.820 & 0.799 & 0.799 \\
& mBERT & 0.760  &   0.753&     0.755     \\
 &XLM-R  & 0.783 &    0.782   &  0.780  \\
 &XLM-R-LD& 0.774   &  0.772  &   0.772  \\
\hline
\end{tabular}
\caption{The scores for individual language varieties on the test set with Naive Bayes, Adaptative Naive Bayes, mBERT, XLM-R, and XLM-R-LD on track two in terms of Recall, Precision, and F1-score. Macro average is reported for average. The best average results are in bold.}
\label{tab:testing2}
\end{table}

\noindent Both the average and individual class results are substantially higher in track two than in track one, once again evidencing the challenge of modeling the `both or neither' class and its impact on the overall performance of the models. In this task formulation, however, track two brings the most important baseline results for DSL-TL as those are obtained when discriminating only between the language varieties. We believe that the `both or neither' is of much less importance to real-world language identification systems.  

\section{Conclusion and Future Research}
\label{conclusion}

This paper presented DSL-TL, the first human-annotated dataset for language variety identification and language identification more broadly. The dataset includes newspaper texts written in three languages and annotated with six language variety labels and a `both or neither' class. We evaluate the performance of multiple models on this dataset, including variations of a classical machine learning approach (Naive Bayes) and multiple deep learning systems (mBERT, XLM-R, and XLM-R-LD). In terms of performance, we observed that the Naive Bayes system delivers performance on par with the deep learning models corroborating the findings of previous research \cite{medvedeva-kroon-plank:2017:VarDial,jauhiainen2019survey}. In certain scenarios, the performance by Naive Bayes even surpassed deep learning models. 

This is a completely new way of looking at the problem. Our findings indicate that there is room for improvement in the treatment and computational modeling of the `both or neither' class. Although this class is of less importance to real-world applications than the variety labels, the low results for this class evidence the challenge of modeling it in this novel language identification setting.

This new dataset opens several avenues for research in language identification. It allows the community to perform a much fairer evaluation of language identification systems mitigating potential biases. We anticipate the true labels strategy presented in DSL-TL to become a new standard in language variety identification, helping to improve the performance of IR and NLP applications that struggle to deal with language variation, such as virtual assistants (e.g., Alexa, Siri), machine translation systems, text and multimedia retrieval systems, and many others.

In the future, we would like to expand the size of this dataset to further investigate the impact of dataset size on performance. We would also like to carry out the same annotation on groups of very similar languages, such as Bosnian, Croatian, and Serbian. DSL-TL is the official dataset of a homonymous ongoing competition at the 2023 edition of the VarDial workshop.\footnote{\url{https://sites.google.com/view/vardial-2023/shared-tasks}} The results presented in this paper will serve as baseline results for the competition.

\section*{Acknowledgements}
We would like to thank the creators of the DSLCC for making the data available. We further thank the VarDial DSL-TL shared task participants for the feedback provided. This work has been partially supported by a GWBC seed fund awarded by RIT, the Academy of Finland (Funding decision no.~341798) and by the Finnish Research Impact Foundation from its Tandem Industry Academia funding in cooperation with Lingsoft.

\bibliography{custom}
\bibliographystyle{acl_natbib}

\end{document}